\documentclass[conference]{IEEEtran}
\IEEEoverridecommandlockouts
\usepackage{cite}
\usepackage{url}
\usepackage{amsmath,amssymb,amsfonts}
\usepackage{algorithmic}
\usepackage{graphicx}
\usepackage{epsfig}
\usepackage{textcomp}
\usepackage{xcolor}
\usepackage{textpos}
\def\BibTeX{{\rm B\kern-.05em{\sc i\kern-.025em b}\kern-.08em
    T\kern-.1667em\lower.7ex\hbox{E}\kern-.125emX}}
\begin{document}

\title{Disclosure of a Neuromorphic Starter Kit}

\author{\IEEEauthorblockN{James S. Plank}
\IEEEauthorblockA{\textit{EECS Department} \\
\textit{University of Tennessee}\\
Knoxville, TN \\
jplank@utk.edu}
\and
\IEEEauthorblockN{Bryson Gullett}
\IEEEauthorblockA{\textit{EECS Department} \\
\textit{University of Tennessee}\\
Knoxville, TN \\
bgullet1@utk.edu}
\and
\IEEEauthorblockN{Adam Z. Foshie}
\IEEEauthorblockA{\textit{EECS Department} \\
\textit{University of Tennessee}\\
Knoxville, TN \\
afoshie@utk.edu}
\and
\IEEEauthorblockN{Garrett S. Rose}
\IEEEauthorblockA{\textit{EECS Department} \\
\textit{University of Tennessee}\\
Knoxville, TN \\
garose@utk.edu}
\and
\IEEEauthorblockN{Catherine D. Schuman}
\IEEEauthorblockA{\textit{EECS Department} \\
\textit{University of Tennessee}\\
Knoxville, TN \\
cschuman@utk.edu}

}
\maketitle

\begin{abstract}
This paper presents a Neuromorphic Starter Kit,
which has been designed to help a variety of research
groups perform research, exploration and real-world
demonstrations
of brain-based, neuromorphic processors and hardware
environments.  A prototype kit has been built and
tested.  We explain the motivation behind the kit,
its design and composition, and a prototype 
physical demonstration.
\end{abstract}


\section{Introduction}

Neuromorphic computing is a computational methodology inspired by the brain.  The
main construct in neuromorphic computing is the spiking neural network (SNN), which
is explained by many sources~\cite{rjp:19:twb,spp:17:sur}.  We use the term
{\em neuroprocessor} to define a computing device on which one may load
a spiking neural network, and then apply input spikes temporally to specific
input neurons.  The neuroprocessor processes the spikes and runs the
SNN, propagating spikes throughout the network.  There are designated output
neurons, where spikes may be read from the outside world.

There are many neuroprocessor simulators~\cite{ks:17:nest,dpd:18:fsd,bbh:14:nap} 
and hardware projects~\cite{fgt:14:tsp,d:18:loihi-long,a:15:tdt-long,spp:17:sur}.  However,
most of the hardware projects are commercial, or run by research programs in ways
that are exclusive to the various research groups.  Our intent with this work
is to provide a low-cost, flexible hardware kit that researchers may use to
explore neuromorphic computing.  In particular, our goal is for the kit to 
enable a straightforward and inexpensive mechanism for developing physical
applications driven by a neuroprocessor.  One of our inspirations is a project
from Delft University, where the authors implemented a neuromorphic PID controller for
adjusting altitude in MAV's~\cite{sdg:22:dip}.  The authors had a clear need for a small, light, 
self-encapsulated system for converting sensor input to spikes, sending those spikes
to a neuroprocessor and then interpreting the output spikes.  We have designed the kit
for uses cases like this one.

In the subsequent sections of this paper, we describe the components of a
kit, their composition, an example kit and a physical application.

\section{Components of a Kit}

\begin{figure}[ht]
\begin{center}
\epsfig{figure=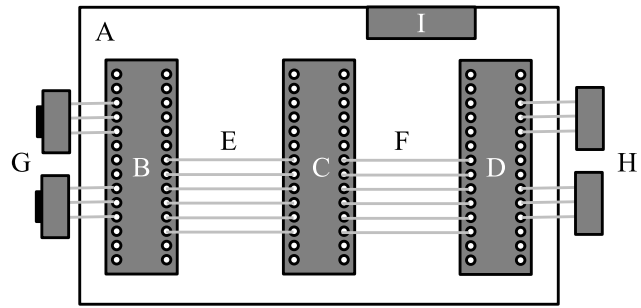,width=3.5in}
\caption{\label{fig:overview} Overview of the components of a Neuromorphic Starter Kit.}
\end{center}
\end{figure}

The basic components of a kit are shown in Figure~\ref{fig:overview}.
Below, we provide detail on the labeled components.

\begin{itemize}
\item {\bf A. The base.}  The base holds components B through F.  For prototype
work, it may be a breadboard or protoboard, or for more hardened work, a printed
circuit board.
\item {\bf B. Input-to-Spike Unit.} This is a component that transforms inputs (G),
often from physical sensors, into spikes that are sent to the neuroprocessor (C) using
the input wires (E).  In our initial work, this is a microcontroller programmed
to interpret the sensor data, and then convert them to spikes, which are either
binary voltage outputs or PWM outputs.  
We provide more detail on this conversion
in Section~\ref{sec:itos} below.
This component can be a more robust
processor like a single-board computer, or an FPGA configured for the specific task.
We selected a microcontroller because of its price, low power usage, speed and
ease of programming.
\item {\bf C. Neuroprocessor.} This is a component that is configured with a SNN,
and receives input spikes from the Input-to-Spike Unit (B) using the input wires (E).
It processes the spikes on the network, and outputs spikes, typically binary
voltage outputs, over the output wires (F) to the Spike-to-Output Unit (D).
In our initial implementation, this component is a microcontroller like units B
and D. However we are designing FPGA's for this component.  Because of their
inherent parallelism, the FPGA's will be able to process spikes at a far greater speed
than the microcontroller.  On the other end of the spectrum, a small single-board computer
(such as the Raspberry Pi Zero) can implement this functionality, albeit with higher complexity
and power requirements.  We provide more detail on the neuroprocessor below in Section~\ref{sec:neuroprocessor}.
\item {\bf D. Spike-to-Output Unit.} This is a component that transforms the output
spikes of the Neuroprocessor (C), which it receives over the output wires (F),
into output values or events that may be interpreted by the outputs, which are
often physical object drivers, like a robotic motor control board.
As with the Input-to-Spike Unit (B), our current kit employs a microcontroller for
this unit, for the same reasons.  We provide more detail on how this component 
converts spikes to outputs in section~\ref{sec:stoo} below.
\item {\bf E. Input Wires.} These are how the Input-To-Spike Unit (B) is connected
to the Neuroprocessor (C).  In our initial prototype, these are standard wires for
a breadboard; however, as a kit becomes more hardened, a printed circuit board may
be designed for the Base (A), and these wires will be configured into the board.
\item {\bf F. Output Wires.} These are how the Neuroprocessor (B) is connected to the
the Spike-to-Output Unit (D).  They are analogous to the Input Wires (E).
\item {\bf G. Inputs.}  These are the inputs to the neuromorphic system.  In our
initial kit, we have one input which is an infrared sensor that is wired to the
inputs of the microcontroller implementing the Input-to-Spike Unit (B).  Multiple
inputs of differing types may be employed as well, and for more complex applications,
we can envision the Inputs to be embodied by a conventional microprocessor connected to
the serial input port of the microcontroller, which then converts the inputs to spikes.
\item {\bf H. Outputs.}  These are the outputs to the neuromorphic system.  In our
initial kit, the output is a motor driver board that controls a robotic car.  There
may be multiple outputs, for example that control individual caterpillar treads of a 
neuromorphically controlled robot~\cite{mbd:17:neon,afd:20:ggr}; or, as with the inputs,
the outputs may be conventional microprocessor (perhaps the same one as G), 
reading serial inputs from its USB port.
\item {\bf I. Power Unit.}  This provides power to the system, and may be attached
to the Base (A) as pictured, or separate from the base, connected by appropriate
wiring.
\end{itemize}

In the following subsections, we provide more detail on components B, C and D.

\subsection{The Input-to-Spike Unit (B)}
\label{sec:itos}

Converting input values or events to spikes for a SNN has been a well-studied research
area~\cite{y:18:fsn,qbg:11:smb,d:21:atn,srm:22:eed,spb:19:nte}.  In our implementation
of the Input-to-Spike Unit, we allow the user to specify, using JSON, a variety of 
input encoding techniques, including population coding (sometimes called ``binning''),
rate coding, spike coding temporal coding, and their combinations.  We build the firmware
for the microcontroller using the PlatformIO embedded systems 
tool (\url{https://docs.platformio.org}).  The JSON is defined by the TENNLab
Exploratory Neuromorphic Computing Framework~\cite{psb:18:ten,spp:21:sfc}.  The JSON
is provided to the PlatformIO compiler, which we have configured to then implement
the corresponding input-to-spike encoding onto the microcontroller.

The spikes produced by the Input-to-Spike Unit may be binary (expressing their values
using bins, spike trains or time-to-first spike), or they may communicate values using
PWM modulations.  As we show in Section~\ref{sec:sizeandspeed}, 
the latter technique can complicate the Neuroprocessor (C), and as 
such, we strive to train SNN's that accept binary spikes as inputs.  

\subsection{The Neuroprocessor}
\label{sec:neuroprocessor}

In our prototype implementation, we have written simulators of two neuroprocessors
developed by the TENNLab research team: RAVENS~\cite{frd:22:bcs} and RISP~\cite{pzg:22:risp}.
Both employ clocked integrate-and-fire neurons, and synapses with integer delays.
Both include programmable neuronal leak, and RAVENS also includes optional STDP on its
synapses, plus configurable refractory periods.
We train SNN's for these neuroprocessors in simulation and store the networks
in JSON.  We then process the JSON with PlatformIO and load the resulting firmware,
which includes both the SNN and the code to process it, onto the microcontroller.

We are developing two RAVENS neuroprocessor implementations on FPGA.  The first 
is like the FPGA neuroprocessors DANNA2~\cite{mdb:18:dda} and Caspian~\cite{msp:20:cnd},
where an SNN is loaded onto the FPGA, which then stores it and runs it.  The second,
which we call ``Strict,'' is like the microcontroller implementation above, where
the SNN is included as input to the Vivado FPGA development tool, and therefore is 
embedded into the programmed FPGA.  As such, the Strict implementation can accomodate
larger networks on the same FPGA.

It would be an interesting project to employ a hardened ASIC such as
Loihi~\cite{d:18:loihi-long} as the neuroprocessor.  This would only employ a subset of
Loihi's functionality, as Loihi's design encompasses all of the functionality of the
kit.  One benefit of the kit would be for neuroprocessor designers to focus solely on
the processing of the SNN, and not worry about input and output processing, thereby
simplifying their processors greatly.

\subsection{The Spike-To-Output Unit (D)}
\label{sec:stoo}

This unit is analogous to the Spike-To-Input Unit, focusing on decoding spikes rather
than encoding them.  As with encoding, there are various techniques for 
decoding~\cite{srm:22:eed}, and our implementation allows the user to specify
them with JSON defined by TENNLab.  The JSON is included as input to the PlatformIO
compiler, and the spike decoding part of the firmware loaded onto the microcontroller.

\section{Training}

The Neuroprocessor assumes that the SNN that it employs has been trained or designed
externally. Then it must be stored as JSON in the format specified by
the TENNLab software framework.  TENNLab networks have been trained using genetic
algorithms~\cite{smp:20:eons,pkc:21:moh,sbk:22:evi}, deep learning~\cite{svd:19:tdn}, 
decision trees~\cite{spp:21:sfc}, and they have been hand-designed for various 
tasks~\cite{pzs:21:snn,pzh:20:rss,hdk:20:mes}.  It is anticipated that in the near-term,
network training will require support from TENNLab software; however, 
that is not a strict requirement.

\section{Prototype Kit}

\begin{figure}[ht]
\begin{center}
\includegraphics[width=3.5in]{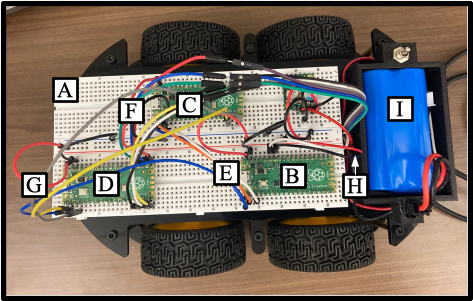}
\caption{\label{fig:top} Annotated view of the top of the prototype Neuromorphic
Starter Kit.}
\end{center}
\end{figure}

\begin{figure}[ht]
\begin{center}
\includegraphics[width=3.5in]{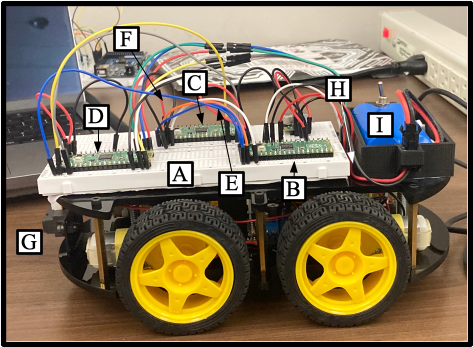}
\caption{\label{fig:side} Annotated view of the top of the prototype Neuromorphic
Starter Kit.}
\end{center}
\end{figure}

We show annotated pictures of our prototype kit, which is used for a robotic car application,
in Figures~\ref{fig:top} and~\ref{fig:side}.

The Base (A) is a standard 1280-point breadboard.
The Input-to-Spike Unit (B), Neuroprocessor (C) and Spike-to-Output Unit (D) are each
implemented with a Raspberry Pi RP2040 Pico microcontroller 
(\url{https://www.adafruit.com/product/4864}).  The input (G) is a Sharp GP2Y0A21YKOF
infrared proximity sensor, and the output (H) and robot uses a Hosyond Smart Robot Car Kit.
Together with wires and various equipment, the total price in summer, 2022 is under \$90.

We wrote an application within the TENNLab software framework to simulate the car,
and hand-designed a simple RISP network whose
objective is for the car to maintain a constant distance from the object in front of
it, while the object randomly moves forward and backward.
The spike encoder converts the infrared sensor reading into a series of spikes
using a ``flip-flop'' encoder~\cite{spb:19:nte}, that has two input neurons for reading
spikes.
There are two output neurons -- one for moving forward and one for moving backward.
The output decoder records the difference between the spike counts in the two output
neurons and converts that into a value that determines the magnitude of moving the robot
forward or backward.

\section{Size and Speed of the Neuroprocessor}
\label{sec:sizeandspeed}

The prototype kit application employs an input encoder that sends spikes as PWM signals
to the Neuroprocessor.  The Neuroprocessor is configured to run at a very slow speed, 10 Hz,
to ease the task of reading the PWM signals.  Configuring an input encoder to produce binary
spikes will allow the Neuroprocessor to speed up drastically to roughly 5 kHz.
Both RISP and RAVENS are implemented on the microcontroller.
In Table~\ref{tab:specs}, we provide estimates of maximum SNN size and estimated Neuroprocessor
speed of four Neuroprocessor implementations: the RP2040 microcontroller, and three FPGA's.
The FPGA estimates are for RAVENs implementations.  We have not implemented RISP on FPGA yet, but anticipate that its simplicity will allow for higher neuron and synapse counts.

\begin{table}[ht]
\begin{center}
\begin{tabular}{l|c|c|c|c}
Neuroprocessor & Price & Neurons & Synapses & Speed \\
\hline
Raspberry Pi RP2040 & \$4 & 512 & 512 & 5 KHz \\
UPduino v3.1 & \$30 & 16 & 256 & 32 MHz \\
CMOD Spartan-7 & \$70 & 40 & 640 & 40 MHz \\
CMOD Artiz-7 & \$100 & 56 & 896 & 52 MHz \\
\end{tabular}
\caption{\label{tab:specs} Neuroprocessor implementations and their estimated price, capacity and speed.}
\end{center}
\end{table}

\section{Near Term Goals}

In the near term, we intend to partner with interested application research groups
to help them develop demonstrations of neuromorphic control applications such as
the prototype.  We are also interested in developing smart sensors that use
the Neuroprocessor to filter sensor data, identifying times when the data should
be forwarded to more conventional computer for processing, but sequestering the
data when it is deemed uninteresting.
These activities will have the benefit of developing several Input-To-Spike
and Spike-To-Output Units for multiple types of inputs and outputs.

\section{Acknowledgements}

This material is based on work supported by
an Air Force Research Laboratory Information Directorate grant (FA8750-19-1-0025).
The authors thank ChaoHui Zheng for writing the application within the
TENNLab framework that simulates the robotic car.

\bibliographystyle{plain}

\end{document}